\documentclass[conference]{IEEEtran}
\usepackage{enumerate}
\usepackage{graphicx}
\usepackage{epsf}
\usepackage{amsmath}
\usepackage{amssymb}
\usepackage{array}
\usepackage{setspace}
\usepackage[amsmath,thmmarks,amsthm]{ntheorem}
\usepackage[ruled,lined]{algorithm2e}
\usepackage{algorithmic}
\usepackage{color}
\usepackage{amsfonts}
\usepackage{dsfont}
\usepackage{xfrac}
\usepackage{breqn}
\usepackage{bbm}
\usepackage{cite}
\usepackage{array}
\usepackage{makecell}
\usepackage[english]{babel}
\usepackage{mathtools}
\usepackage{mathrsfs}
\usepackage[labelformat=simple]{subcaption}

\graphicspath{{fig/}}

\theoremseparator{.}

\begin{document}
\title{\Huge Edge Selection and Clustering for Federated Learning in Optical Inter-LEO Satellite Constellation} 


\author{\IEEEauthorblockN{Chih-Yu Chen, Li-Hsiang Shen, Kai-Ten Feng, Lie-Liang Yang\textsuperscript{*}, and Jen-Ming Wu\textsuperscript{\textdagger}}\\
 \IEEEauthorblockA{Institute of Communications Engineering, 
  National Yang Ming Chiao Tung University, Hsinchu, Taiwan\\ 
  \textsuperscript{*}Next Generation Wireless, University of Southampton, Southampton, UK\\
  \textsuperscript{\textdagger}Next-generation Communications Research Center, Hon Hai Research Institute, Taipei, Taiwan\\
  leochen0904.ee08@nycu.edu.tw,
  gp3xu4vu6.cm04g@nctu.edu.tw, 
  ktfeng@nycu.edu.tw, \\
  lly@ecs.soton.ac.uk,
  and jen-ming.wu@foxconn.com
  }}

\maketitle

\begin{abstract}
     Low-Earth orbit (LEO) satellites have been prosperously deployed for various Earth observation missions due to its  capability of collecting a large amount of image or sensor data.
     However, traditionally, the data training process is performed in the terrestrial cloud server, which leads to a high transmission overhead.
     With the recent development of LEO, it is more imperative to provide ultra-dense LEO constellation with enhanced on-board computation capability. 
     Benefited from it, we have proposed a collaborative federated learning for low Earth orbit (FELLO). 
     We allocate the entire process on LEOs with low payload inter-satellite transmissions, whilst the low-delay terrestrial gateway server (GS) only takes care for initial signal controlling. 
     The GS initially selects an LEO server, whereas its LEO clients are all determined by clustering mechanism and communication capability through the optical inter-satellite links (ISLs). The re-clustering of changing LEO server will be executed once with low communication quality of FELLO.
     In the simulations, we have numerically analyzed the proposed FELLO under practical Walker-based LEO constellation configurations along with MNIST training dataset for classification mission. 
     The proposed FELLO outperforms the conventional centralized and distributed architectures with higher classification accuracy as well as comparably lower latency of joint communication and computing.
\end{abstract}

\section{Introduction} \label{sec:INTRO}
    With the advancement in space technology and the increasing demand on non-terrestrial services, the development of satellite systems, especially low-Earth orbit (LEO) satellites, has attracted much attention in research recently \cite{Sat_ovw0, LEO_appl0, 6G_ovw0}.
    The industry and organization, e.g., SpaceX, OneWeb and Kuiper have invested great efforts in LEO constellations thanks to its lower cost, shorter distance to ground and resilient networking  \cite{CON_ovw1}.
    The needs in applications such as image capturing, weather or geographic data collection are explosively growing. 
    Huge amount of data keeps being collected by the moving LEOs, which has driven the utilization of artificial intelligence (AI) for processing these resources.
    
    Traditionally, limited by the stringent computation power and memory on satellites, the data processing tasks are performed at ground gateway server (GS).
    All data collected in the space needs to be transmitted back to the ground, which results in high transmission delay and controlling overhead.
    However, with recent development on inter-satellite links (ISLs) \cite{ISL_ovw0, ISL_ovw1, ISL_ovw2}, ISLs can build up connections and collaboration among satellites in a more efficient and effective way. 
    It helps improve joint communication and computing among LEOs and relieve the communication overhead with the GS.
    However, fast moving LEOs may break the LEO-GS link due to the misaligned beamforming direction.
    Also, LEOs have only limited contact time to a GS. 
    The potential client idleness and model staleness during the training process will  increase the training time \cite{FedSpace}, \cite{FedISL}.
    
    Benefited from low-cost deployment, dense LEOs forming mega-constellations are empowered with on-board computation capability \cite{OEC_ovw0}, \cite{OEC_ovw1}. 
    LEOs will be capable of providing more complex and globally-covered services, or even collaborative training tasks.
    This opens up an opportunity toward federated learning (FL)\cite{FL, FL_ovw2}.
    With the aid of FL, LEOs can process the data locally and only exchange the encrypted model among LEOs through a candidate edge server, which reduces the transmission data size and the time induced from long distance between LEOs and the GS.
    Recent works have been discussing the benefits and challenges to establish FL on satellite systems\cite{FL_ovw0, FL_ovw1}. 
    In \cite{FedSat}, the authors improve the asynchronous FL with the scheduling algorithm in LEO constellation. 
    In \cite{FedSpace}, the authors propose the FL algorithm with local model buffer and scheduling to balance the trade-off between satellite idleness and model staleness.
    The paper \cite{FedISL} tries to alleviate staleness by multi-hop ISL to frequently send back FL model parameters to the GS. 
    The authors in \cite{FedHAP} further introduce the high-altitude platform system as edge.
    The authors in \cite{AFLS} have proposed an asynchronous two-layer aggregation FL, where the first layer is on LEOs and the second aggregation is on the GS. 
    However, papers of \cite{FL_ovw0, FL_ovw1, FedSat, FedSpace, FedISL, FedHAP, AFLS} require model or data feedback to the GS for computation, which leads to comparatively higher latency due to long distance between LEOs and GS.
    
    To the best of our knowledge, we are the first work to propose federated learning for low Earth orbit (FELLO) consisting of only operating LEO constellation.
    Note that the GS is only responsible for initial signal controlling with considerably lower overhead compared to raw data processing, FELLO considers all data process on the orbit, which has well-addressed the bottleneck of LEO-GS transmission delay overhead. 
    Due to high mobility of LEOs, constellations will change rapidly over time.
    Therefore, to tackle potential expired model as well as low link quality, we propose an LEO edge selection and clustering (LESC) scheme to cluster the candidate neighboring LEOs as FL clients.
    We evaluate the performance of FELLO under mega-constellation of around 700 satellites under practical constellation setting. 
    Simulation results show that the proposed LESC in FELLO outperforms the conventional centralized and distributed learning architectures in terms of accuracy and training overhead. 
    
    The rest of the paper is outlined as follows. 
    In Section \ref{sec:SYS}, we introduce the system model, signaling and LEO constellation of FELLO.
    In Section \ref{sec:PROS}, we describe the proposed LESC algorithm in FELLO. 
    In Section \ref{sec:SR}, we demonstrate the performance evaluation of FELLO.
    Finally, conclusions are drawn in Section \ref{sec:CON}.
    
\section{System Architecture and Model} \label{sec:SYS}
\subsection{Architecture and Signalling of FELLO} \label{sub:SYSFL}
    
    \begin{figure}[t]
        \centerline{\includegraphics[width=8cm]{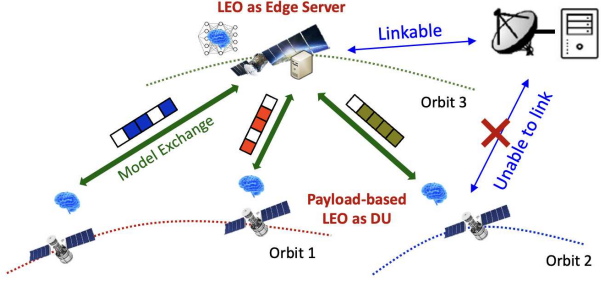}}
        \caption{System model for FELLO.}
        \label{fig:sys1}
    \end{figure}
    
\begin{figure}[htbp]
        \centerline{\includegraphics[width=8cm]{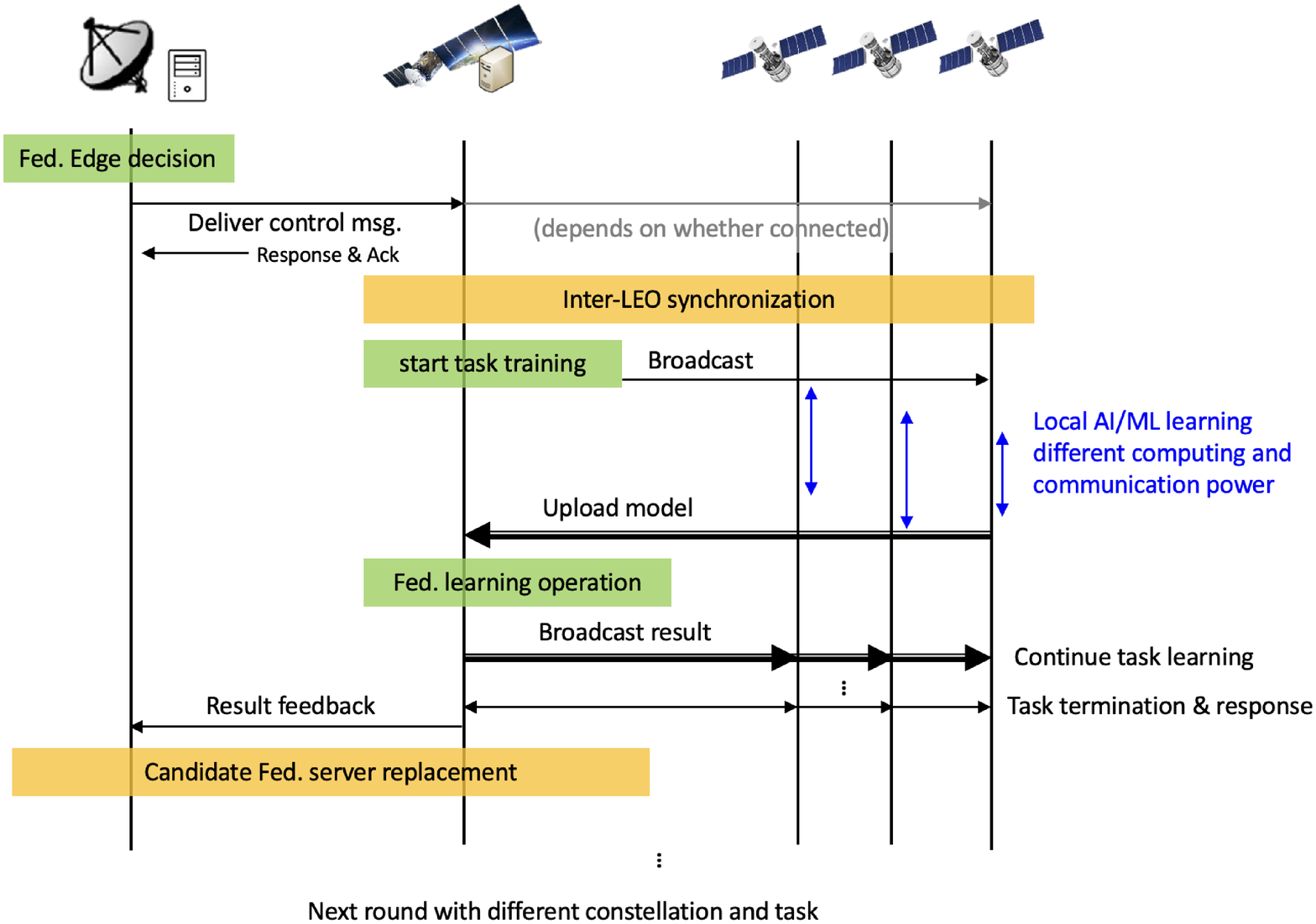}}
        \caption{The signalling of the proposed FELLO system.}
        \label{fig:sys2}
    \end{figure}

    The system architecture of proposed federated learning over LEO constellation (FELLO) is shown in Fig. \ref{fig:sys1}.  The whole system consists of three parts, including the terrestrial GS, LEO groups of a single edge and several client nodes. 
    The entire process is described in Fig. \ref{fig:sys2}. 
    The GS first determines an LEO satellite with better communication link quality as an FL edge server and transmits the control signal with the assigned training tasks. 
    The edge server then selects a cluster of LEO clients $\mathcal{K} = \{1,2,\cdots,K\}$ as FL clients with feasible link. 
    Each LEO client acquires data on its on-board equipment, and trains the model locally without sharing the raw data to either the GS or other LEOs due to high transmission overhead. 
    After several data collection and task training rounds, the clients within the FL cluster will transmit the compressed trained model to the LEO edge. 
    The edge server will then aggregate the received model and send back the updated parameters to the associated LEO clients. 
    Due to high mobility of LEOs, the connected links of ISL over LEOs might lead to failure after certain session.
    The LEO edge server will handover the task to the upcoming reachable candidate LEO to the GS, whilst the new LEO edge server will re-cluster the clients to continue the task.

\subsection{FL Model}
    We adopt the concept of FedAvg \cite{FedAvg} in our system, which employs an average manner for model aggregation.
    LEO client $k$ has local dataset $S_k$ with $N_k$ data points, where each data point consists of $x_{n}$ as input data vector and $y_{n}$ as the ground truth label, respectively.
    We define the local loss function for each LEO client $k$ given by 
    \begin{equation}
        \label{eq:localL}
        L_k(w) = \frac{1}{N_k} \sum^{N_k}_{n = 1} l_n(x_n, y_n, w_n),
    \end{equation}
    where $l_n(x_n, y_n, w_n)$ is the loss function of data sample $(x_n, y_n)$ and model parameter is defined as $w_n$. 
    The global loss function is therefore given by
    \begin{equation}
        \label{eq:globalL}
        L(w) = \frac{1}{N} \sum^K_{k=1} \sum_{n=1}^{N_{k}} l_n(x_n, y_n, w_n),
    \end{equation}
    where $N = \sum^K_{k = 1} N_k$ is the total data samples.
    The objective of the FL is to find the optimized model parameter $w^*$ that minimizes the global loss function as
    \begin{equation}
        \label{eq:localmin}
        w^* = \arg \min_{w} L(w).
    \end{equation}
    For every aggregation round $a \in \{ 1, 2, \cdots, A \}$, LEO edge server will transmit the global model $w_G^a$ to every client. 
    Each client $k$ will perform local gradient descent to train its local parameter $w_k^a$. 
    For each FL aggregation round, clients will update the parameters for $E$ local training epochs. For each epoch $e$, the local model parameter $w_{k}^{a, e}$ is given by
    \begin{equation}
        \label{eq:localgd}
        w_k^{a, e} = w_k^{a, e-1} - \eta g_{k}^{a, e-1},
    \end{equation}
    where $\eta$ is the learning rate and $g_{k}^{a, e} = \nabla_{w_k} L_k(w_k^{a, e})$ is the local gradient.
    After the local update, the client will transmit the weight $w_k^a = w_k^{a, E}$  to the FL edge server. The server will perform the aggregation to generate the global model parameter for next FL round as 
    \begin{equation}
        \label{eq:globalagg}
        w_G^{a+1} = \sum^{K}_{k=1} \frac{N_k}{N} w_k^a.
    \end{equation}

\subsection{Optical ISL Channel and Signal Model} \label{sub:SYSISL}
    We consider laser communication for ISL with the received signal power given by 
    \begin{equation}
        \label{eq:PR}
        P_R = P_T \eta_T \eta_R G_T G_R L_T L_R L_{PS} ,
    \end{equation}
    where $P_T$ is transmission power, $\eta_T$ and $\eta_R$ are efficiency constants of respectively optical transmitter and receiver, $G_T$ and $G_R$ are transmitter and receiver antenna gains, $L_T$ and $L_R$ are transmitter and receiver pointing loss factor, and $L_{PS}$ stands for the path loss \cite{ISL1}. 
    The transmitter and receiver are assumed to use the same telescope with the same antenna gain as
    \begin{equation}
        \label{eq:GTGR}
        G_T = G_R = \left( \frac{\pi D}{\lambda} \right) ^2, 
    \end{equation}
    where $D$ is the telescope diameter, and $\lambda$ is the signal wavelength.
    The laser misalignment loss can be expressed as 
    \begin{equation}
        \label{eq:LT}
        L_{x} = \exp\left(-G_{x}\theta_{x}^2\right), \forall x \in \{T, R\},
    \end{equation}
     where $\theta_T$ and $\theta_R$ are respectively the transmitter and receiver radial pointing error angles. From \cite{ISL1}, \cite{ISL0}, the probability distribution function of radial pointing error angle ($\theta \in \{\theta_T, \theta_R\}$) can be described by a Rayleigh distribution function 
    \begin{equation}
        \label{eq:theta}
        f(\theta) = \frac{\theta}{\sigma_\theta} \exp \left(-\frac{\theta^2}{2\sigma_\theta ^2} \right),
    \end{equation}
    where $\sigma_\theta$ is standard deviation (SD) of $\theta$.
    The path loss in ISL is considered as free space transmission given by 
    \begin{equation}
        \label{eq:LPS}
        L_{PS} = \left(\frac{\lambda}{4 \pi d} \right)^2 ,
    \end{equation}
    where $d$ is the distance between transmitting and receiving LEO satellites.
    

    
    Moreover, the photon detector and amplifier modules in laser communications bring additional noise to the signal. 
    We consider the signal shot noise, dark current noise and thermal noise, which are respectively defined as
    \begin{equation}
        \label{eq:ssn}
        \sigma_{sn}^2 = 2 q R_p P_R B,
    \end{equation}
    \begin{equation}
        \label{eq:sdc}
        \sigma_{dc}^2 = 2 q I_d B,
    \end{equation}
    \begin{equation}
        \label{eq:sth}
        \sigma_{th}^2 = \frac{4 k_B T B}{R_L},
    \end{equation}
    where $q$ is the electron charge, $R_p$ is the responsivity, $B$ is the bandwidth of LEO, $I_d$ is the dark current, $k_B$ is the Boltzmann constant, $T$ is noise temperature, and $R_L$ is the load resistance \cite{ISL2}. 
    According to \eqref{eq:ssn}-\eqref{eq:sth}, the total noise power is obtained as
    \begin{equation}
        \label{eq:PN}
        P_N = \sigma_{sn}^2 + \sigma_{dc}^2 + \sigma_{th}^2.
    \end{equation}
    The signal-to-noise ratio (SNR) for ISL is therefore defined as 
    \begin{equation}
        \label{eq:SNR}
        \gamma = P_R / P_N.
    \end{equation}
    Based on \eqref{eq:SNR}, we can acquire the achievable rate as
    \begin{equation}
        \label{eq:R}
        R = \left(1-p_e\right) B \cdot \log_2(1 + \gamma),
    \end{equation}
    where $p_e$ indicates the bit error rate (BER) relevant to different modulation and coding schemes. 
    Note that we neglect interference in laser communications due to  its considerably narrow laser beams than radio-based beamforming.
    
\subsection{LEO Mega-Constellation} \label{sub:SYSCON}
    We consider a widely-adopted Walker constellation as the orbit shell \cite{CON1}, which provides global coverage under the uniform deployment of orbit planes. 
    Based on different inclination configurations, Walker constellations can be categorized into two basic types, including Walker Star and Walker Delta. 
    Walker Star possesses near-polar orbits with orbit inclination close to 90\textdegree.
    On the other hand, Walker Delta focuses on services for middle and low-latitude regions where most population resides. 
    From \cite{CON0}, the position of LEO $k$ in orbit plane $l$ in Cartesian coordinate $P_{l,k}(x_{l,k}, y_{l,k}, z_{l,k})$ is formulated as 
    
    \begin{equation}
        \begin{split}
        \label{eq:coX}
        x_{l, k}(t) = R_S [\cos(\Omega_l(t)) \cos(\omega_{l, k}(t)) - \\ \sin(\Omega_l(t)) \sin(\omega_{l, k}(t)) \cos(\phi)],
        \end{split}
    \end{equation}
    \begin{equation}
        \begin{split}
        \label{eq:coY}
        y_{l, k}(t) = R_S [\sin(\Omega_l(t)) \cos(\omega_{l, k}(t)) - \\ \cos(\Omega_l(t)) \sin(\omega_{l, k}(t)) \cos(\phi)],
        \end{split}
    \end{equation}
    \begin{equation}
        \label{eq:coZ}
        z_{l, k}(t) = R_S \sin(\omega_{l, k}(t)) \sin(\phi), 
    \end{equation}
    where $R_S$ is the radius of the orbit which is equal to the sum of Earth radius $R_E$ and the altitude of LEO $h_S$, i.e., $R_S = R_E + h_S$, and $\phi$ is the constant orbit inclination.
    $\Omega_l(t)$ and $\omega_{l, k}(t)$ denote the inter-plane and intra-plane angular positions of orbit plane $l$ and LEO $k$ at time instant $t$, which are respectively given by
    \begin{equation}
        \label{eq:Omgl}
        \Omega_l(t) = \Omega_0(l) + \dot\Omega t ,
    \end{equation}
    and 
    \begin{equation}
        \label{eq:omglm}
        \omega_{l,k}(t) = \omega_0(l,k) +\dot\omega t ,
    \end{equation}
    where $\dot\Omega = 7.292115856 \times 10^{-5}$ rad/s is the Earth rotation rate and $\dot\omega = \frac{\sqrt{398601.2}}{R_S^{3/2}}$ rad/s is the orbit velocity.
    Also, $\Omega_0(l)$ and $\omega_0(l,k)$ are both used to describe the initial position. 
    For Walker Delta constellation, the initial state is given by
    \begin{equation}
        \label{eq:wdomg}
        \omega_{D0}(l, k) = (k-1) \frac{\pi}{N_S} + (l-1) \frac{\pi}{N_S N_O},
    \end{equation}
    and 
    \begin{equation}
        \label{eq:wdOmg}
        \Omega_{D0} (l) = (l-1) \frac{\pi}{N_O} ,
    \end{equation}
    respectively, where $N_O$ is the number of orbit planes, and $N_S$ is the number of LEOs per orbit plane.
    
\section{Proposed LEO Edge Selection and Clustering (LESC) in FELLO} \label{sec:PROS}

    FELLO system contains two main procedures, including LEO edge selection and client clustering as well as generic federated task learning. The concrete algorithm is elaborated in Algorithm \ref{alg:cluster}.
    Considering a practical case, a GS on the ground has limited time to contact with the LEOs. 
    Therefore, we consider the GS acts as an initiator for selecting an appropriate edge of the whole training process with the best ground-satellite link (GSL) quality $\gamma_{gsl, k}^* = \max \gamma_{gs, k}$, where $\gamma_{gs, k}$ indicates the SNR between the GS and the $k$-th LEO. 
    Afterwards, the edge will cluster a group of adjacent LEO satellites based on the given threshold $\delta_x$, i.e., distance threshold $\delta_d$ or SNR threshold $\delta_\gamma$, as its training client set $\mathcal{K}_a$ in the $a$-th FL aggregation round. 
    For each aggregation round $a$, LEO client $k \in \mathcal{K}_a$ will collect their individual task data, train the neural network model locally, and then transmit the final trained model $w_k^a$ based on \eqref{eq:localgd} to the LEO edge. Afterwards, the LEO edge server will aggregate all the received model weights in an average manner to update the global model $w_G^{a+1}$ via \eqref{eq:globalagg}.
    Once the LEO edge server is unable to be associated with the GS, the GS will assign a new candidate of LEO edge server with sustainable SNR. 
    The original LEO edge server will then handover the task and current global model to the upcoming one with the following training.
    Also, for every aggregation round, LEO server will remove those clients whose link quality cannot meet the threshold $\delta_x$, i.e., either $d < \delta_d$ or $\gamma > \delta_\gamma$. 
    For every re-clustering period of $T_{rc}$ aggregation rounds, once the number of participating LEOs is smaller than the given threshold $\epsilon$, i.e., $K_a < \epsilon K'$, where $K'$ is the temporarily recorded cluster size, the LEO edge server will re-cluster a group of clients based on $\delta_x$.
    With the threshold of $\delta_d$, the LEO edge server clusters all clients in a convex shape. 
    On the contrary, due to the fluctuating channel, the transmission quality differs over times with irregular non-convex shape under the SNR threshold of $\delta_\gamma$, which guarantees ISL quality. 

    \begin{algorithm}[!tb]
    \caption{Proposed LEO Edge Selection and Clustering (LESC) in FELLO}
    \SetAlgoLined
    \DontPrintSemicolon
    \label{alg:cluster}
    \begin{algorithmic}[1]
    
    \STATE Initialization: LEO constellation, SNR threshold for LEO-GS link $\gamma_{th}$ , number of FEL aggregation rounds\\ $A$, re-clustering period $T_{rc}$, LEO client cluster $\mathcal{K}_a$ for round $a$ indexed by $k$ with size $K_a$, the clustering threshold $\delta_x, \forall{x} \in \{d, \gamma\}$, the re-clustering threshold $\epsilon$ 
    	\FOR{$a=1,2,\dots,A$}
    	\IF{$a=1$ or LEO-GS SNR $\gamma_{gsl, k}^* < \gamma_{th}$}
    		\STATE GS selects the nearest LEO as edge server
    		\STATE LEO edge server clusters LEO clients with link quality meeting the threshold $\delta_x$ as $\mathcal{K}_1$ with size $K_1$
            \STATE Set initial cluster size $K' = K_1$
    	\ENDIF
    	\IF{$a>1$}
    		\STATE LEO edge server removes LEO clients $k$ with low link quality, i.e., $\gamma < \delta_\gamma$, or $d > \delta_d$, and then $\mathcal{K}_a \leftarrow \mathcal{K}_{a-1} - k$
    		\IF{$K_a < \epsilon K'$ and $\mod\left(a, T_{rc}\right) = 0$}
    			\STATE LEO edge server re-clusters LEO clients $\mathcal{K}_a$ with size $K_a$ and update $K' = K_a$
    		\ENDIF
    	\ENDIF
            \FOR{\textbf{each} client $k$ in $\mathcal{K}_a$ \textbf{in parallel}}
                \STATE Conduct local task training for $e=\{1,...,E\}$ \\epochs based on \eqref{eq:localL} and \eqref{eq:localgd}
                \STATE Obtain the final local model $w_{k}^{a}$
                \STATE Upload model $w_{k}^a$ to LEO edge server via ISL
            \ENDFOR
    		
    		\STATE LEO edge server performs FedAvg based on \eqref{eq:globalagg} as $w_G^{a+1} = \sum_{k\in \mathcal{K}_a} \frac{n_k}{N} w_{k}^a$
    		\STATE LEO edge server broadcasts the global model $w_G^{a+1}$ to  LEO clients $\mathcal{K}_{a+1}$ as their next learning model
    	\ENDFOR
    
    \end{algorithmic}
    \end{algorithm}

\section{Performance Evaluation} \label{sec:SR}
    We demonstrate the performance of proposed LESC in FELLO system.
    Simulations are conducted under a single-shell based Walker Delta constellation \cite{CON1}, referring to as one of the orbit shell in Starlink's multi-layer constellation. 
    The parameters of the orbit shell and the ISL links are shown in Table \ref{table:param}. 
    We employ image dataset of MNIST for classification task using a 3-layer multi-layer perception (MLP) model. 
    The entire training process for federated learning is conducted over 40 FL aggregation rounds. 
    Every time the global LEO edge server performs FL aggregation, whilst the LEO clients are supposed to train $E = 2$ epochs.
    Two benchmark architectures are considered for comparison, including centralized and distributed learning (CL/DL).
    Note that the training process in CL is conducted with the same running epochs and data size. 
    Compared to FELLO, the main difference in CL is that all data collected from each client will be transmitted to a chosen centralized LEO server performing CL method. 
    On the other hand, the LEO clients in DL conduct local training without any information exchange over ISL.

    \begin{table}[htbp]
    \centering
    \footnotesize
      \caption{Parameters Setting}
     \begin{tabular}{l*{1}{l}r}
      \hline
      $\textbf{Parameter}$ & $\textbf{Value}$ \\
      \hline
      Constellation  & Walker Delta \\
      Number of LEOs & 720 \\
      Number of orbits & 36 \\
      Inclination & $70$\textdegree \\
      Altitude ($h_S$) & $570$ km \\
      \hline
      Laser wavelength ($\lambda$) & $1500$ nm \\
      Bandwidth ($B$) & $1.25$ GHz \\
      Transmit power ($P_T$) &  $30$ mWatt\\
      Transmit optical efficiency ($\eta_T$) & $0.8$ \\
      Receiving optical efficiency ($\eta_R$) & $0.8$ \\    
      Receiving telescope diameter ($D_R$) & $60$ mm \\
      Pointing error angle SD ($\sigma_\theta$) & $3\ \mu$rad\\
      Responsivity ($R_p$) & 0.6007 \\
      Dark current ($I_d$) & $1$ nA\\
      Noise temperature ($T$) & $500$ K \\
      Load resistance ($R_L$) & $1000$ Ohm\\
      \hline
      Total data size ($N$) & 60000 \\
      Number of data per client ($N_k$) & 2208 \\
      Total FL rounds ($A$) & 40 \\
      Local training epochs ($E$) & 2 \\
      Batch size & 32 \\
      Re-cluster threshold of LEOs ($\epsilon$) & 0.7 \\  
      SNR threshold for LEO-GS link ($\gamma_{th}$) & 20 dB \\
      \hline
     \end{tabular} \label{table:param}
    \end{table}

\subsection{Comparison with Benchmark Architecture}
    
\begin{figure}[ht]
    \centering
    \begin{subfigure}{0.245\textwidth}
        \centering
        \includegraphics[width=\textwidth]{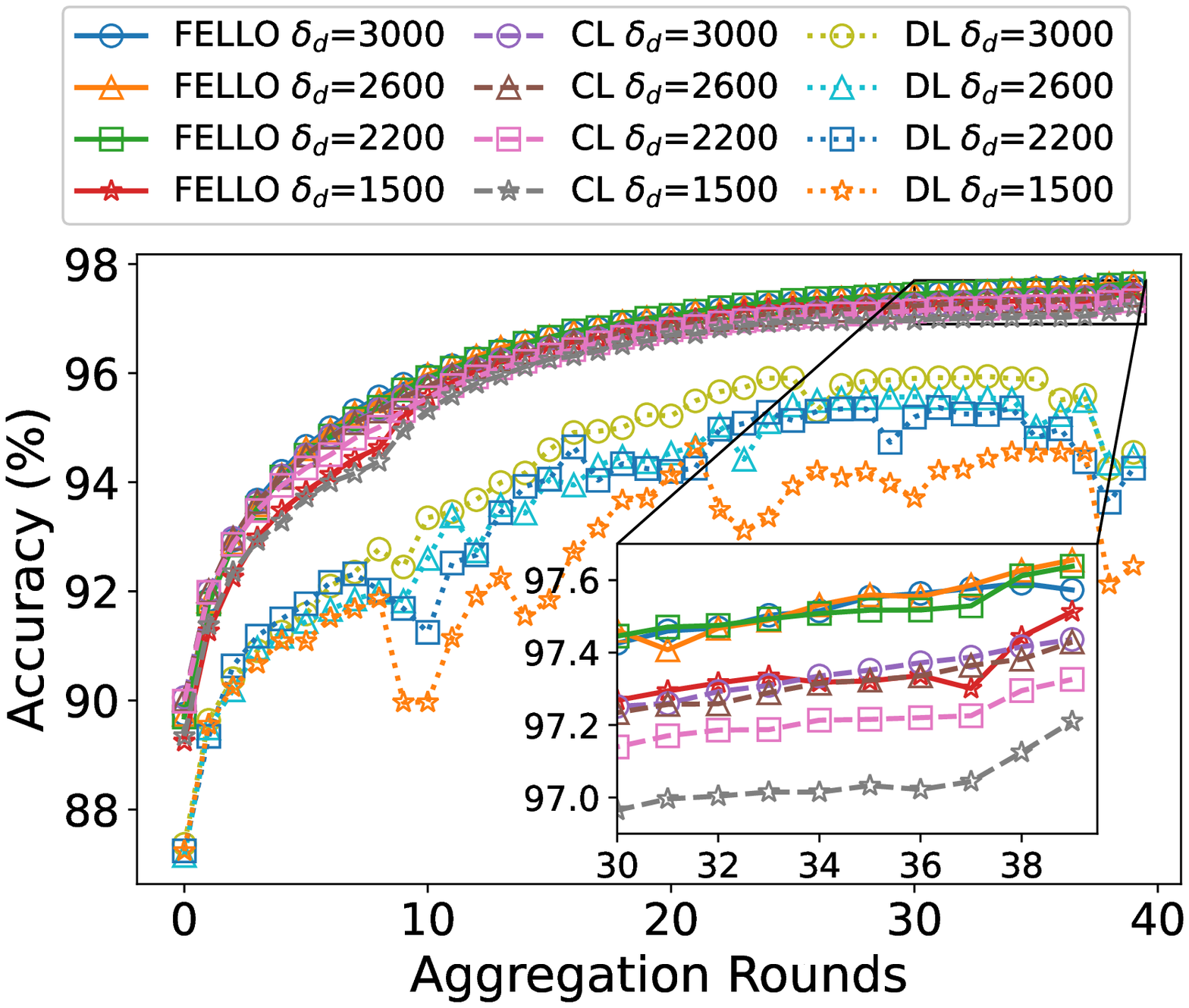} 
        \caption{}
        \label{fig:R_r}
    \end{subfigure}
    \hspace{-0.8em}
    \begin{subfigure}{0.245\textwidth}
        \centering
        \includegraphics[width=\textwidth]{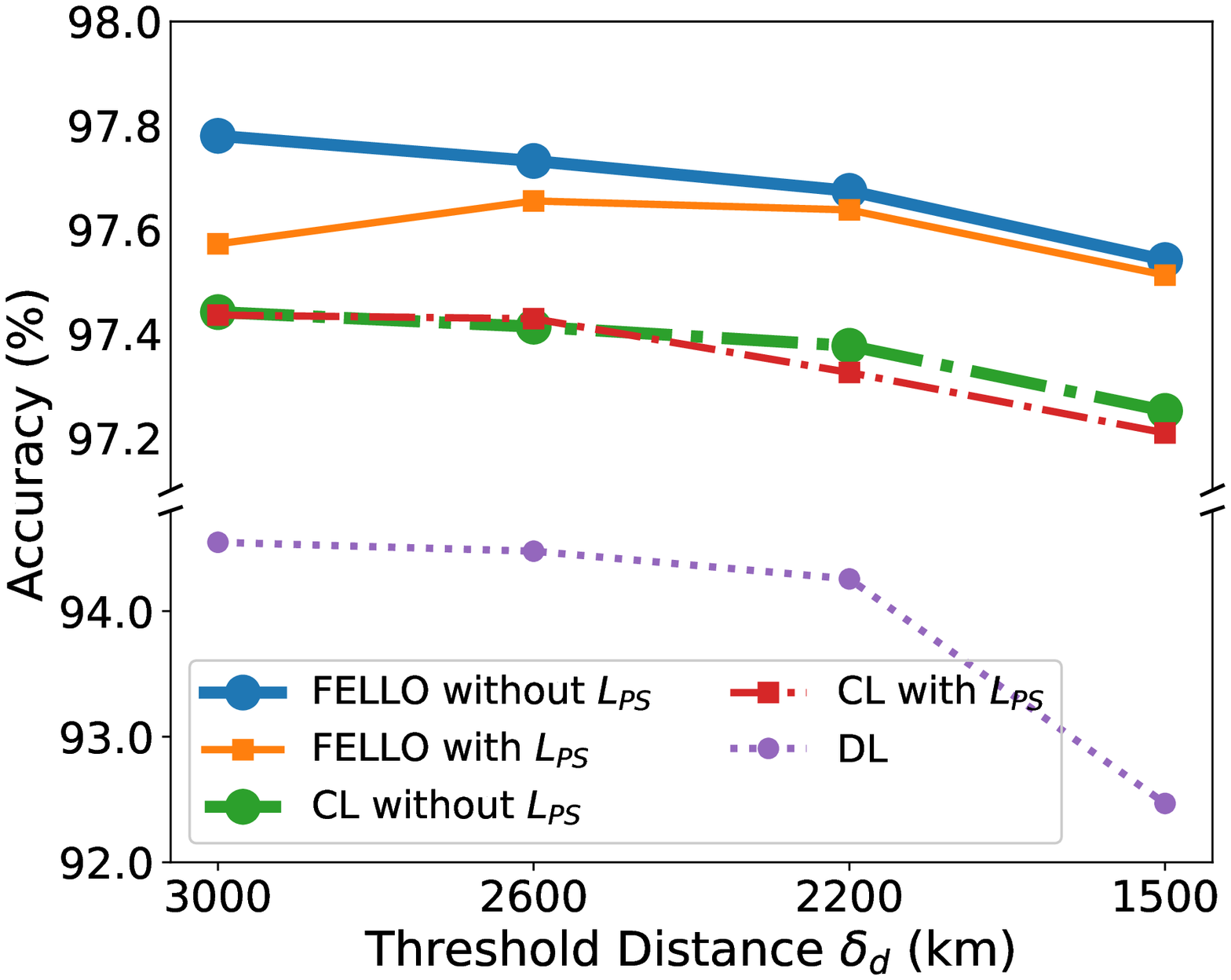}
        \caption{}
        \label{fig:R_m}
    \end{subfigure}
    \caption{Performance comparison of proposed FELLO system and CL/DL architectures in terms of (a) convergence and (b) accuracy with different distance thresholds of $\delta_d$.}
    \label{fig:R}
    \end{figure}

    In Fig. \ref{fig:R}, we evaluate the effect of different distance thresholds of $\delta_d \in \{ 1500, 2200, 2600, 3000\}$ km, which allow the LEO edge server to cluster around $\{5, 12, 18, 24\}$ LEO clients. 
    We can observed that FELLO outperforms both CL and DL with the highest accuracy. 
    With more clients involved, the model converges faster and result in higher classification accuracy. 
    However, higher $\delta_d$ with larger cluster size will include clients with larger path loss involved. 
    This induces an optimal solution under $\delta_d=2600$ km with imperfect model exchange due to either erroneous packets from long distances or fewer participants from short ranges. 
    To elaborate a little further, DL has the worst accuracy due to no information exchange among other LEOs.

    \begin{figure}[ht]
    \centering
    \begin{subfigure}{0.245\textwidth}
        \centering
        \includegraphics[width=\textwidth]{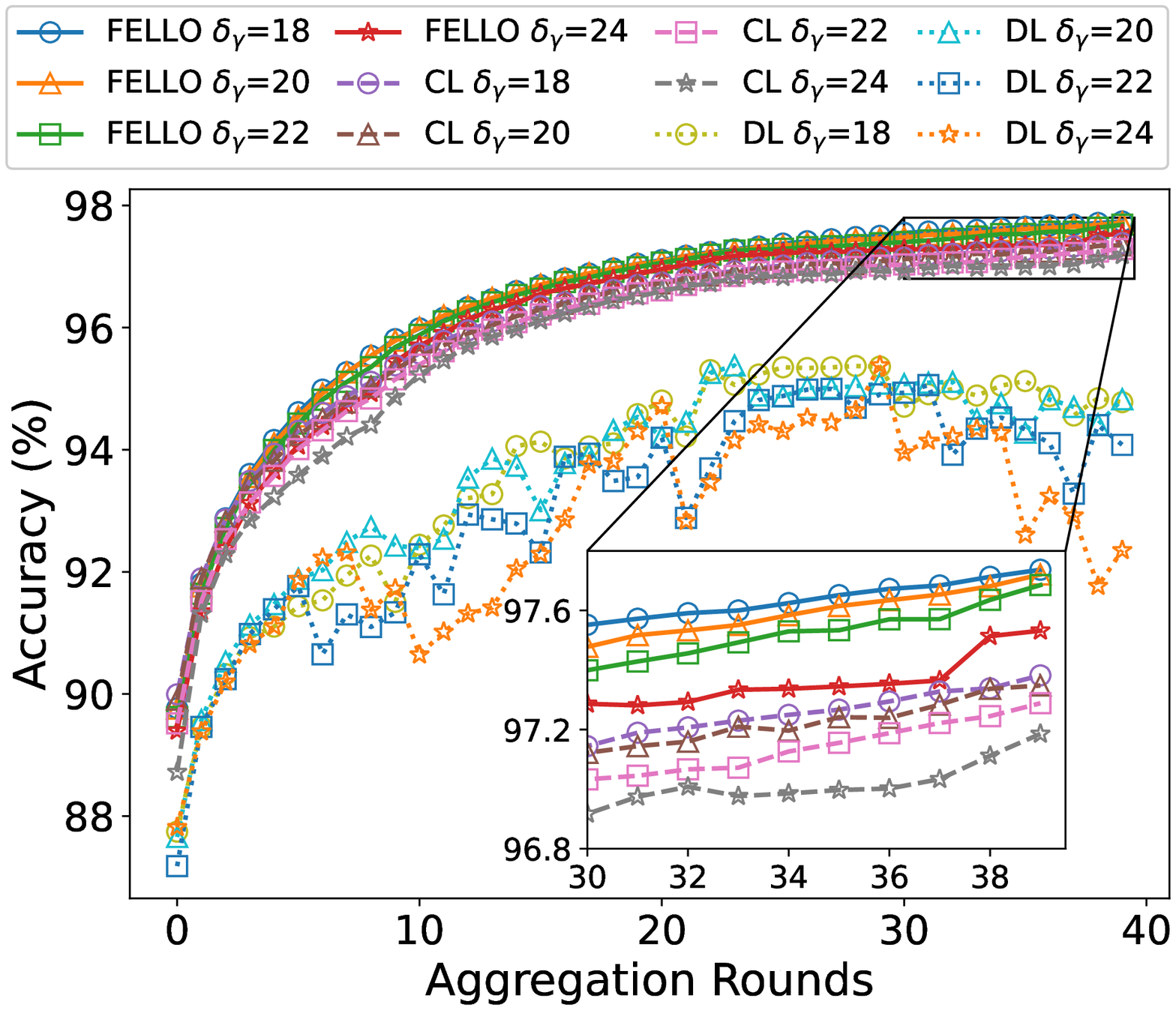} 
        \caption{}
        \label{fig:SNR_r}
    \end{subfigure}
    \hspace{-0.8em}
    \begin{subfigure}{0.245\textwidth}
        \centering
        \includegraphics[width=\textwidth]{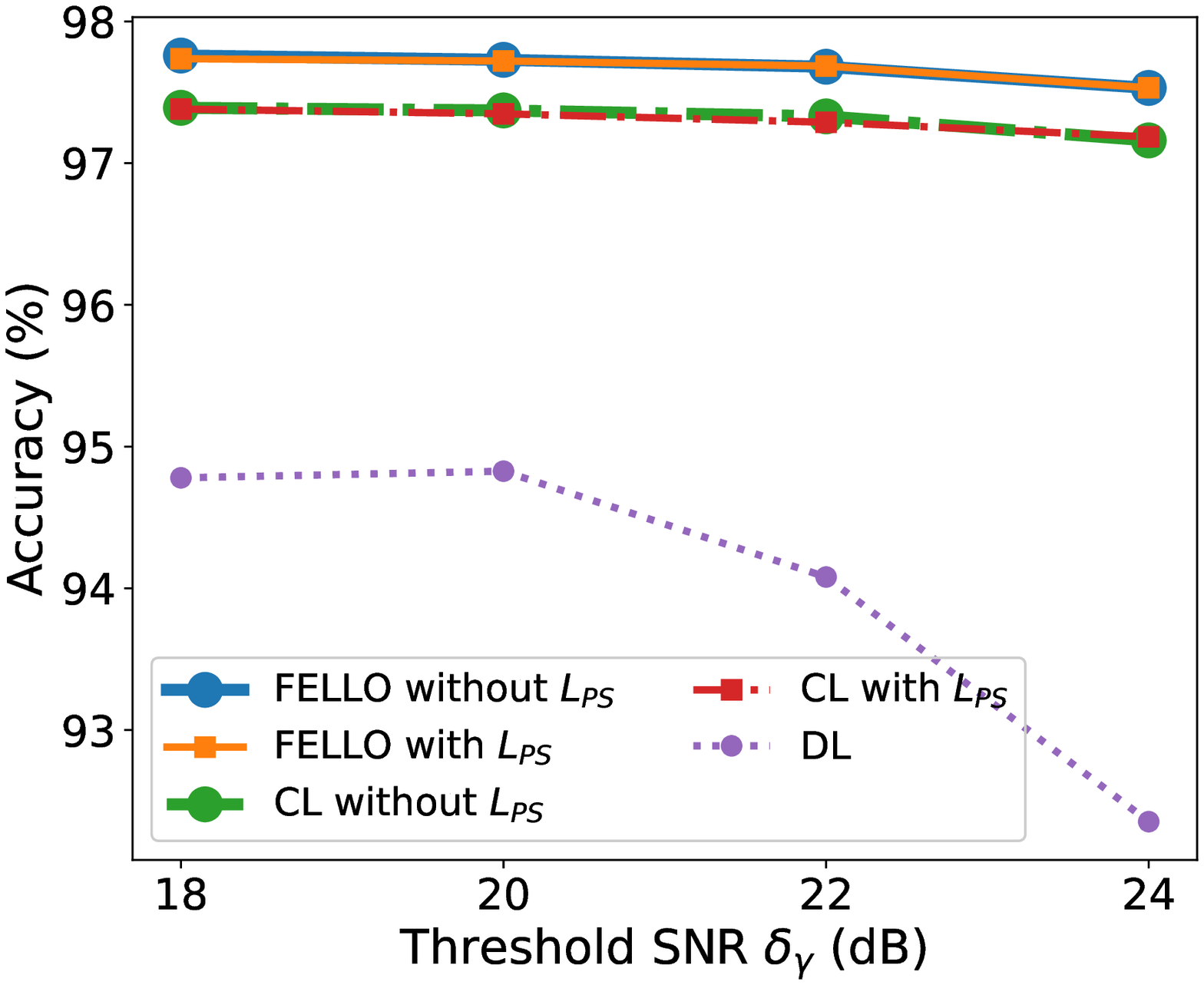}
        \caption{}
        \label{fig:SNR_m}
    \end{subfigure}
    \caption{Performance comparison of proposed FELLO system and CL/DL architectures in terms of (a) convergence and (b) accuracy with different SNR thresholds of $\delta_\gamma$.}
    \label{fig:SNR}
    \end{figure}

    In Fig. \ref{fig:SNR}, we cluster clients by the SNR thresholds of  $\delta_\gamma \in \{18, 20, 22, 24\}$ dB. 
    With higher SNR threshold $\delta_\gamma$, FELLO will cluster LEO clients with comparatively smaller path loss. 
    Therefore, the classification accuracy with path loss applied on transmitted model asymptotically approaches to the ideal condition without path loss. 
    However, the link quality of ISL may vary due to channel randomness and satellite mobility, which results in larger variation on clustering size than that of distance threshold based method.
    This potentially degrades the training results and sometimes causes divergence.

    \begin{figure}[ht]
    \centering
    \begin{subfigure}{0.245\textwidth}
        \centering
        \includegraphics[width=\textwidth]{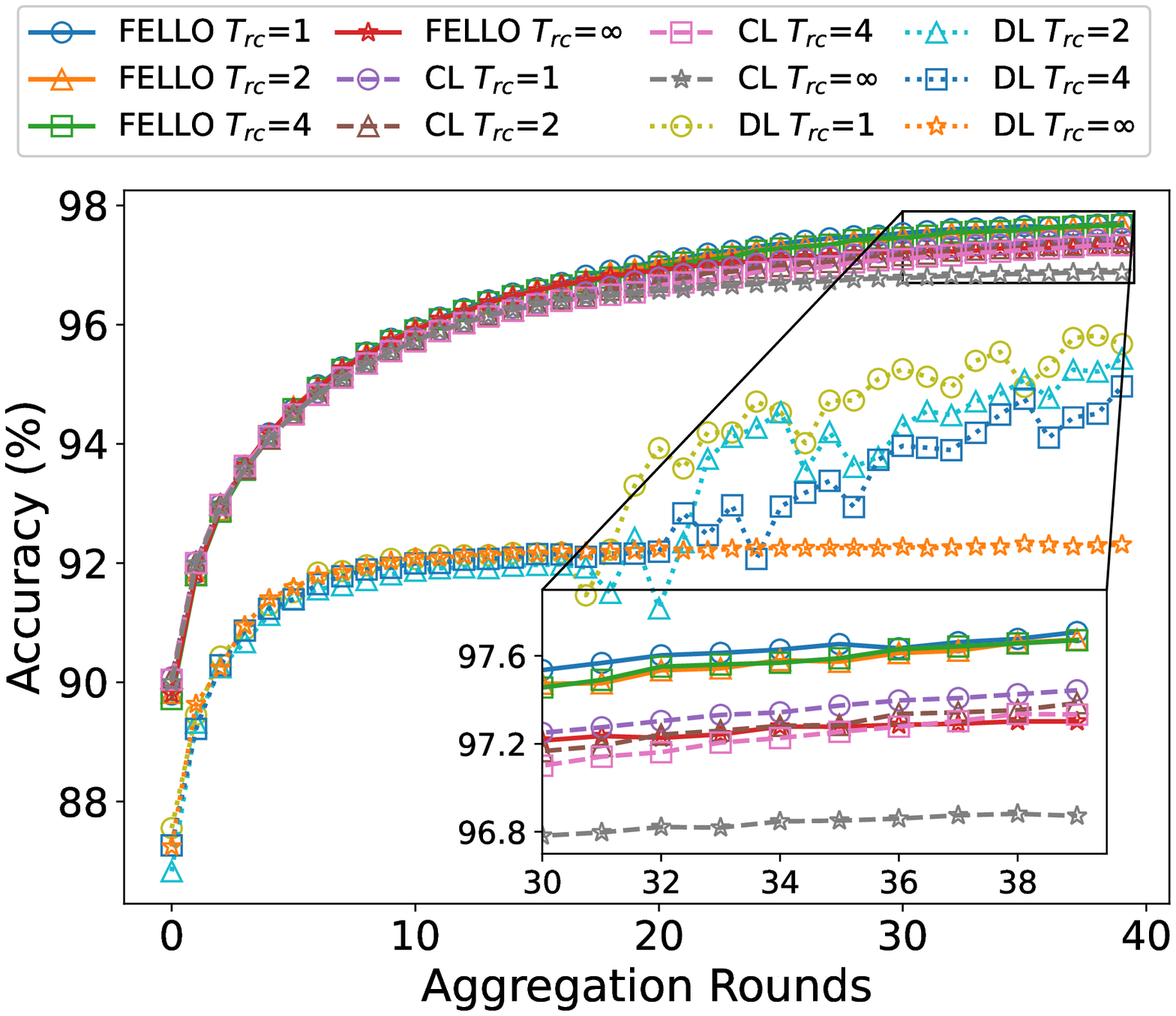} 
        \caption{}
        \label{fig:Trc_r}
    \end{subfigure}
    \hspace{-0.8em}
    \begin{subfigure}{0.245\textwidth}
        \centering
        \includegraphics[width=\textwidth]{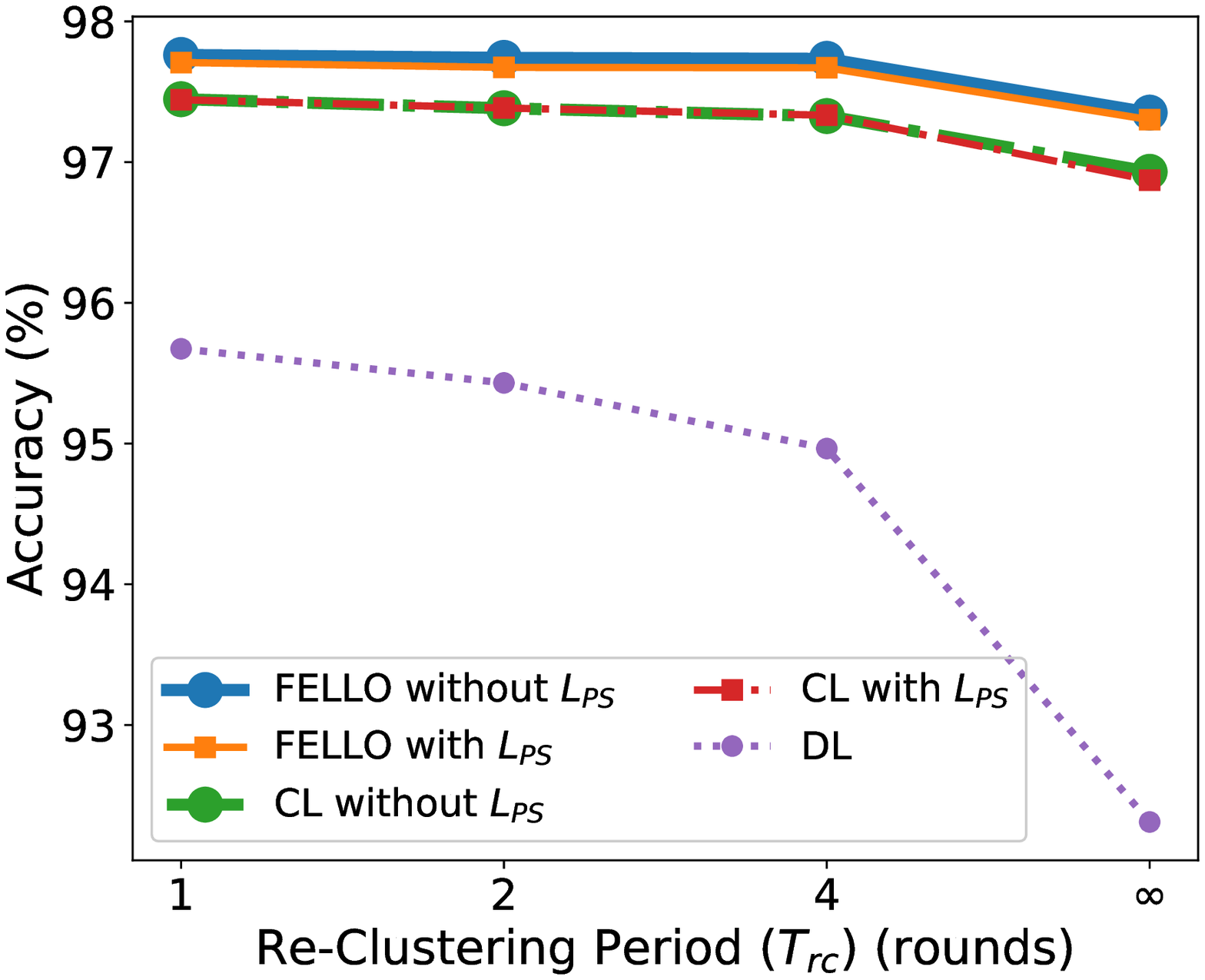}
        \caption{}
        \label{fig:Trc_m}
    \end{subfigure}
    \caption{Performance comparison of proposed FELLO system and CL/DL architectures in terms of (a) convergence and (b) accuracy with different re-clustering periods of $T_{rc}$.}
    \label{fig:Trc}
    \end{figure}

    Furthermore, the re-clustering is a key mechanism for FELLO to deal with the time-varying channel. 
    The simulation under $\delta_d = 2600$ km is shown in Fig. \ref{fig:Trc}. 
    We select the re-clustering period as $T_{rc} \in \{1, 2, 4, \infty\}$, where smaller $T_{rc}$ indicates more frequent re-clustering, whilst $T_{rc}=\infty$ means no re-clustering is performed during the whole process. 
    The result shows that if without re-clustering, the classification accuracy of both FELLO and CL degrades significantly. 
    By contrast, more frequent re-clustering with smaller $T_{rc}$ provides better performance, which shows the necessity of re-clustering mechanism in FELLO.

    \begin{figure}[ht]
    \centering
    \begin{subfigure}{0.245\textwidth}
        \centering
        \includegraphics[width=\textwidth]{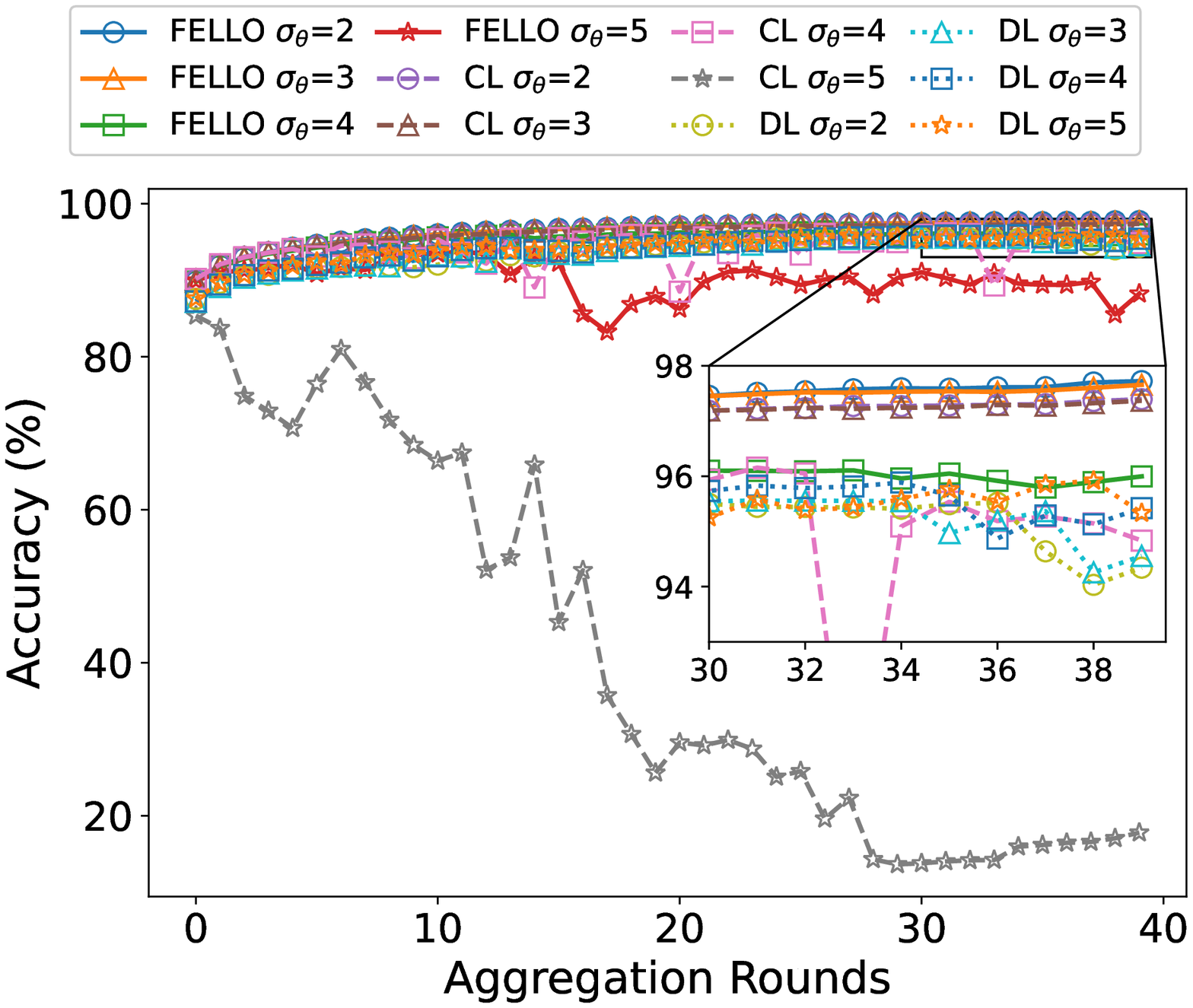} 
        \caption{}
        \label{fig:ps_r}
    \end{subfigure}
    \hspace{-0.8em}
    \begin{subfigure}{0.245\textwidth}
        \centering
        \includegraphics[width=\textwidth]{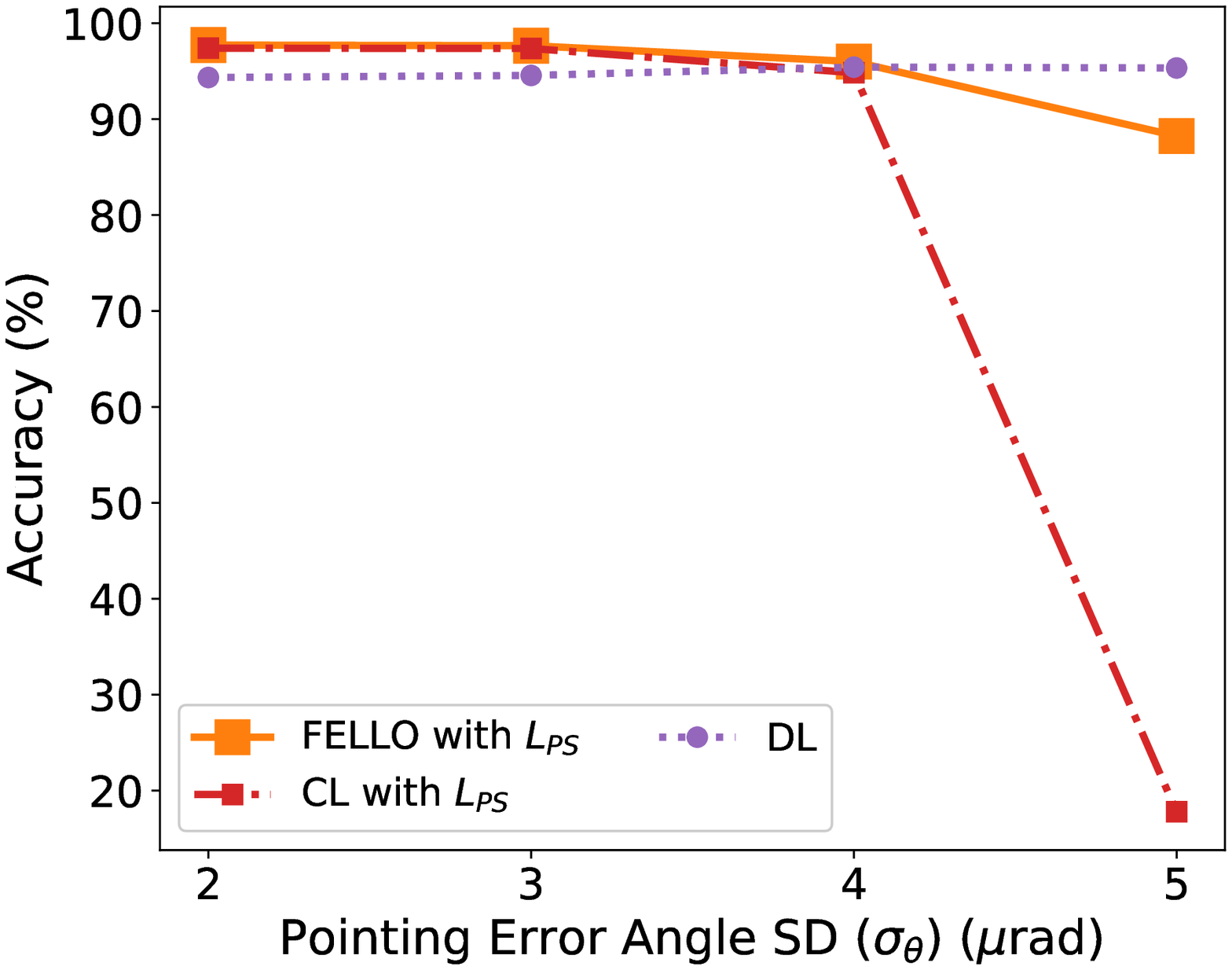}
        \caption{}
        \label{fig:ps_m}
    \end{subfigure}
    \caption{Performance comparison of proposed FELLO system and CL/DL architectures in terms of (a) convergence and (b) accuracy with different pointing error angle SD of $\sigma_\theta$.}
    \label{fig:ps}
    \end{figure}

    To further evaluate the system resilience under harsh channel conditions, we consider different $\sigma_\theta\in\{2,3,4,5\} $ $\mu$rad for misaligning beams in the optical ISL channel in Figs. \ref{fig:ps_r} and \ref{fig:ps_m} with $\delta_d =2600$ km.
    Larger $\sigma_\theta$ causes more severe pointing error on the transmitted models, leading to larger variation on cluster size and more frequent re-clustering with lower accuracy.
    The classification accuracy of DL remains unchanged due to no information exchange between LEOs. 
    
    However, the accuracy of FELLO outperforms CL, benefited by re-clustering of appropriate candidates of LEO clients.
    Under larger variantion of pointing error, CL is more severely affected by channel resulting in  divergence.
    Although DL has stable performance under varying $\sigma_\theta$, it is affected by the time-varying clustering size and cannot converge in most cases.

\subsection{LEO Overhead Analysis}

    Additionally, we analyze the communication and computation overhead in Table \ref{table:FCCOM}. 
    The delay time $T_{tr, x}$ accounts for transmission time over ISL $T_{s, x}$, training time per epoch $T_{e, x}$ and FL parameter aggregation time $T_{a, x}$, where $x\in\{F, C, D\}$ indicates FELLO, CL and DL systems, respectively. 
    The total system delay $T_{tr, F}$ of the proposed FELLO can be derived as
    \begin{equation}
        \label{eq:TFL}
        T_{tr,F} = A \cdot (2T_{s,F} + E \cdot T_{e,F} + T_{a,F}), 
    \end{equation}
    where $T_{e,F}$ is the time for LEO clients to train the local model per epoch, and $T_{a,F}$ is that for the LEO edge server to aggregate local models.
    Additionally, doubled $T_{s,F}$ indicates model upload and download over ISL. Similarly, the delay overhead of DL is asymptotically obtained based on \eqref{eq:TFL} as 
    \begin{align}
        \label{eq:TDL}
        T_{tr,D} = \left(A \cdot E\right) \cdot T_{e,D} .
    \end{align}
    In \eqref{eq:TDL}, we can observe that $T_{e, D}$ is identical to $T_{e, F}$, and there is no model transmission and aggregation time included in total system delay since in DL, clients only condcut local training without information exchange among LEOs. 
    The overhead of CL can be expressed as
    \begin{align}
        \label{eq:TCL}
        T_{tr,C} = T_{s,C} + (A \cdot E) \cdot T_{e,C},
    \end{align}
    where $T_{s,C}$ stands for the time of raw data transfer, whilst $T_{e,C}$ is required to perform comparably longer training than either FELLO or DL due to higher data amount in CL.
    The result is shown in Table \ref{table:FCCOM} with 20 LEO clients in the cluster for fair comparison. 
    The proposed FELLO system requires total delay time of 2.36 s, which has comparably lower delay than CL of 15.67 s. 
    The bottleneck of CL comes from training and transmission of raw data. 
    Also, the memory required for LEO edge server in FELLO is much smaller than that in CL.
    Though with a asymptotic overhead with FELLO, the non-cooperative architecture of DL degrades the accuracy performance without moderate information exchange.
    
    \newcolumntype{L}[1]{>{\raggedright\let\newline\\\arraybackslash\hspace{0pt}}m{#1}}
    \newcolumntype{C}[1]{>{\centering\let\newline\\\arraybackslash\hspace{0pt}}m{#1}}
    \newcolumntype{R}[1]{>{\raggedleft\let\newline\\\arraybackslash\hspace{0pt}}m{#1}}

    \begin{table}[htbp]
    \centering
    \footnotesize
      \caption{System Overhead Analysis}
     \begin{tabular}{L{8em} R{6em} R{6em} R{6em}}
      \hline
      \multicolumn{1}{c}{$\textbf{Parameters}$} & \multicolumn{3}{c}{$\textbf{LEO learning types}$} \\
      & $\textbf{FELLO}$ & $\textbf{Centralized}$ & $\textbf{Distributed}$\\
      \hline
      Architecture & LEOs as  edge/clients& LEOs controlled by GS& LEOs as clients\\
      \hline
      Computation overhead & 0.878 T FLOPS & 17.56 T FLOPS & 0.878 T FLOPS \\
      \hline
      Transmission time ($T_{s, x}$)  & 0.101 ms & 0.445 ms & - \\
      \hline
      Training time per epoch ($T_{e, x}$) & $29.38$ ms & 195.88 ms & 29.38 ms \\
      \hline
      FL aggregation time ($T_{a, x}$) & 0.089 ms & - & - \\
      \hline
      Total system delay ($T_{tr, x}$) & 2.36 s & 15.67 s & 2.35 s \\
      \hline
      Server memory required & 0.52 MB &  140.28 MB & - \\
      \hline
      Client memory required & 7.04 MB &  7.01 MB & 7.04 MB \\
      \hline
     \end{tabular} \label{table:FCCOM}
     
    \end{table}
    
\section{Conclusion} \label{sec:CON}
    In this paper, we conceive an FELLO system, which enables computing in space without comparably high delay overhead of processing at GS owing to long distance. 
    In FELLO, LEOs communicate with each other under optical ISLs, which significantly reduces the communication overhead.
    Also, the LESC algorithm is proposed to select an appropriate LEO edge serve as well as cluster the neighboring LEO clients with decent channel quality.
    We have evaluated the proposed FELLO with LESC in terms of various parameters of clustering thresholds, re-clustering periods, and misalignment effects. 
    Benefited from both ISL and FL, the results have shown that FELLO with LESC outperforms the benchmarks of conventional centralized and distributed LEO architectures in terms of the highest classification accuracy as well as the lowest overhead of total system delay and required memory footprint.

\bibliographystyle{IEEEtran}
\bibliography{IEEEabrv}
\end{document}